\documentclass{article}
\PassOptionsToPackage{numbers,square,sort&compress}{natbib}
\usepackage{iclr2027_conference}
\iclrfinalcopy
\setcitestyle{numbers,square,comma}
\usepackage[T1]{fontenc}
\usepackage[utf8]{inputenc}
\usepackage{amsmath,amssymb,amsfonts}
\usepackage{newtxtext}
\usepackage[varvw]{newtxmath}
\usepackage{algorithmic}
\usepackage{graphicx}
\usepackage{booktabs}
\usepackage{textcomp}
\usepackage{microtype}
\usepackage[font=small,labelfont=bf,labelsep=period]{caption}
\usepackage[hidelinks]{hyperref}
\usepackage{bm}

\newcommand{\etal}{\textit{et al.}}
\newcommand{\blfootnote}[1]{\begingroup\renewcommand{\thefootnote}{}\footnote{#1}\addtocounter{footnote}{-1}\endgroup}
\newcommand{\keywords}[1]{\begin{quote}\small\textbf{Keywords:}\enspace #1\end{quote}}
\makeatletter
\def\@maketitle{\vbox{\hsize\textwidth\centering
  \hrule height 1pt\vskip 0.16in
  {\fontsize{17.28}{21}\selectfont\bfseries\@title\par}
  \vskip 0.16in\hrule height 1pt\vskip 0.18in
  {\normalsize\normalfont\@author\par}
  \vskip 0.02in}}
\makeatother

\begin{document}
\title{Rethinking Cross-Sensor Domain Adaptation for Bridge Point Cloud Semantic Segmentation}
\author{%
Linghui Huang\textsuperscript{1}, \
Zipeng Wang\textsuperscript{2}, \
Shunjie Chen\textsuperscript{1}, \
Peng Guan\textsuperscript{1},\\
Jiandong Wang\textsuperscript{3}, \
Siqing Li, \ and \
Luming Yang\\[6pt]
{\footnotesize\textsuperscript{1}Guangdong Provincial Government Loan Repayment Expressway Management Center, Guangdong, China\\
\textsuperscript{2}China Railway Major Bridge Engineering Group Co., Ltd., Wuhan, China\\
\textsuperscript{3}Guangdong Xixun Intelligent Technology Co., Ltd., Guangdong, China}}
\date{}

\maketitle
\blfootnote{This work has been submitted to the IEEE for possible publication. Copyright may be transferred without notice, after which this version may no longer be accessible.}
\blfootnote{Correspondence: skylynfluke@gmail.com}

\begin{abstract}
Segmentation models for bridge point clouds lose accuracy when trained on terrestrial laser scanning (TLS) data and applied to mobile laser scanning (MLS) data.
This study re-examines the SemanticBridge TLS-to-MLS protocol to determine how sampling density, surface roughness, color, and acquisition viewpoint are expressed in the 8-cm sparse-voxel representation, and how adaptation gains depend on source-model convergence.
Using one sparse-convolution U-Net, we combine raw-point statistics with controlled input perturbations and compare adaptive batch normalization, entropy minimization, Mean Teacher, and confidence-filtered self-training under 100- and 300-epoch schedules.
TLS and MLS have similar nearest-neighbor spacing after preprocessing (12.08 and 12.02~mm), and a 4-mm perturbation of TLS points changes voxel occupancy by only 4.4\%.
Color distributions differ systematically, but zeroing color on identical voxels changes target mean intersection over union (mIoU) by only $+0.02$, and by $+0.06$ after the color augmentation is scale-corrected.
The sensors also observe different portions of the bridge, yielding different class proportions.
A 300-epoch source-only model obtains 57.80 mIoU on the paired-bridge protocol.
Confidence-filtered self-training improves the 100-epoch baseline, but from a converged model it changes target mIoU by $-0.97$ (source-selected checkpoints) and $-2.44$ (final-50-epoch means).
Repeated evaluation (standard deviation 0.27--0.64) and seed variation (1.88 across three seeds) are comparable to several method differences.
Adaptation results on this benchmark therefore depend on representation scale, source-model convergence, and evaluation variability.
\end{abstract}


\section{Introduction}
\begin{figure}[!tbp]
\centering
\includegraphics[width=\textwidth]{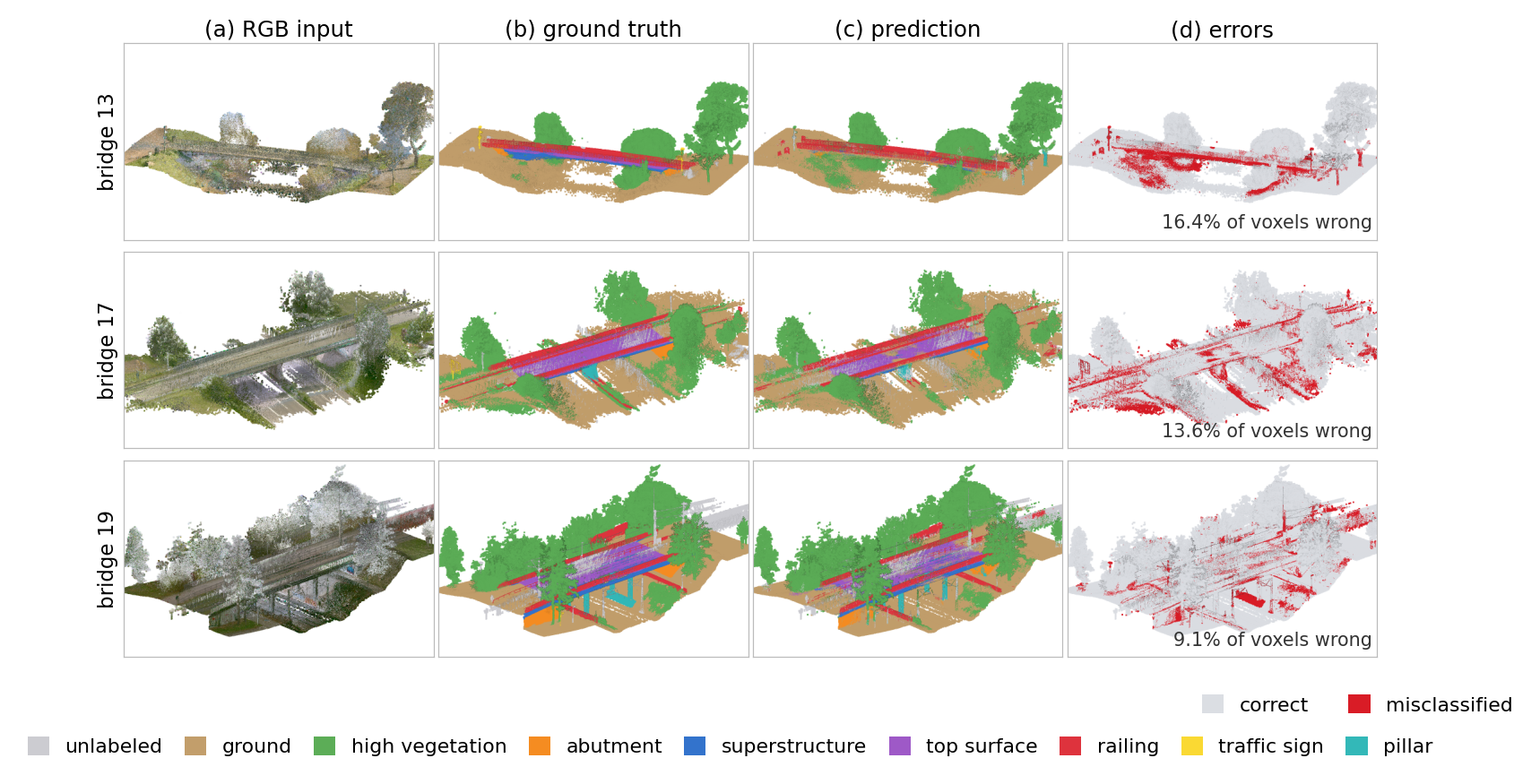}
\caption{Whole-bridge semantic-segmentation predictions for the three target MLS evaluation bridges. The examples preview the practical TLS-to-MLS transfer setting studied in this work: most bridge structures are recovered, while errors concentrate near component boundaries, vegetation-occluded abutments, and unreliable regions of bridge 19.}
\label{fig:qualitative}
\end{figure}

Point cloud semantic segmentation is a prerequisite for automated inspection, geometric measurement, and bridge information modeling \cite{lu2019digital,mafipour2023automated} because downstream systems require point-level separation of decks, abutments, superstructures, railings, and piers.
The same bridge may be scanned at different times by stationary terrestrial laser scanning (TLS), handheld or vehicle-mounted mobile laser scanning (MLS), or light detection and ranging (LiDAR) mounted on unmanned aerial vehicles (UAVs) \cite{lee2026uav}.
UAV platforms also support complementary image-based inspection tasks, including resource-efficient crack recognition under degraded imaging conditions \cite{li2026robust}.
Each platform produces different visibility, noise, and appearance characteristics.
Collecting dense labels for every platform and project is generally uneconomical, which motivates transferring models between sensing systems.
Figure~\ref{fig:qualitative} shows predictions on the three MLS evaluation bridges after training with TLS data.

Earlier bridge-analysis pipelines relied on geometric rules, slicing, and structural priors \cite{lu2019detection,lu2019digital,yang2024domain}, whereas recent methods increasingly use learned point-cloud representations \cite{kim2020automated,lee2021semantic,lin2025structure}.
Most reported systems, however, assume that training and test data are acquired by the same sensing platform.
Outside the bridge domain, sensor and platform changes also reduce segmentation accuracy, for example between automotive LiDAR sensors \cite{langer2020domain}, between airborne datasets \cite{xie2021crosscity}, and from TLS to MLS \cite{shen2023segtrans}, and are addressed by point-cloud domain adaptation \cite{triess2021survey,xiao2024survey}.
SemanticBridge \cite{kellner2026semanticbridge} enables controlled study of this problem by providing TLS and MLS acquisitions of the same bridges under a predefined benchmark split.
Its authors reported TLS-to-MLS losses of 6.9--11.4 mean intersection over union (mIoU) points across three backbones.
Methods developed for automotive LiDAR often assume that beam pattern and density dominate the shift \cite{yi2021complete,wei2022lidar}; appearance-alignment methods assume a consequential photometric mismatch \cite{jaritz2020xmuda,xiao2022transfer}; and test-time adaptation methods assume that low-order feature statistics or output confidence provide an adaptation signal \cite{li2018adabn,wang2021tent}.
This study determines which TLS--MLS differences in density, roughness, color, and viewpoint persist in the network input and how adaptation gains depend on source-model convergence.

We analyze the shift at the input resolution of the segmentation model, namely discrete 8-cm voxel coordinates with red-green-blue (RGB) features.
Raw-point statistics are considered together with controlled perturbations of the model input.
At this resolution, the dataset's 1-cm preprocessing largely equalizes sampling density, the measured roughness difference has a limited effect on voxel occupancy, and paired color interventions change target mIoU by less than 0.1 point with both the benchmark and the scale-corrected color augmentation.
The two sensors also observe different bridge components, which changes their class proportions.

We then evaluate adaptive batch normalization (AdaBN), test entropy minimization (TENT), Mean Teacher, and confidence-filtered self-training under the same data split and under two training schedules.
The comparison also reveals a substantial dependence on source-model convergence: confidence-filtered self-training improves a short-schedule baseline but not a model trained with the longer schedule.
We therefore report learning-rate schedule effects, repeated evaluations, and variation across random seeds alongside the adaptation results.

This paper re-examines the SemanticBridge TLS-to-MLS protocol with one sparse-convolution U-Net; its contributions are as follows.
\begin{itemize}
    \item We compare TLS and MLS at both raw-point and voxel-input levels, considering sampling density, roughness, color, and viewpoint-dependent coverage.
    \item We evaluate AdaBN, TENT, Mean Teacher, and confidence-filtered self-training under a common experimental protocol and two training schedules.
    \item We establish a source-only sparse-voxel baseline that attains 57.80\% MLS mIoU on the benchmark three-pair protocol and analyze the effect of baseline convergence on the estimated adaptation gain.
    \item We measure evaluation variability, seed variation, cross-annotation agreement, and registration quality to provide context for numerical differences between methods.
\end{itemize}

\section{Related Work}
\label{sec:related}
\subsection{Bridge Point-Cloud Segmentation}
Rule-based bridge reconstruction exploits the regular arrangement of structural elements.
Lu \etal\ separated deck and pier groups using top-down slicing, surface normals, and oriented bounding boxes \cite{lu2019detection}, and subsequently converted labeled clusters into information-model components \cite{lu2019digital}.
Yang \etal\ incorporated finite-element-derived structural knowledge into region growing \cite{yang2024domain}, while Xia \etal\ combined local descriptors with conventional machine learning \cite{xia2021automated}.
Such methods are interpretable but sensitive to bridge type and acquisition quality, and their rules and thresholds are designed for data from a single acquisition setup.

Deep models instead learn point-wise labels end to end \cite{guo2021deep,xie2020linking}.
Bridge-specific work includes component recognition \cite{kim2020automated}, hierarchical representations \cite{lee2021semantic} and structure-oriented losses \cite{lin2025structure}, while commonly adopted backbones originate from general 3-D segmentation: PointNet++ \cite{qi2017pointnetpp}, kernel point convolution (KPConv) \cite{thomas2019kpconv}, sparse convolutional networks \cite{choy20194d,graham2018submanifold}, Point Transformer V2 (PTv2) \cite{wu2022ptv2}, and RandLA-Net \cite{hu2020randla}.
Many of these architectures and training recipes were developed on automotive sequences such as SemanticKITTI \cite{behley2019semantickitti} or large static scans such as Semantic3D \cite{hackel2017semantic3d}, whose acquisition geometry differs from bridge surveys.
SemanticBridge evaluates UNet3D, KPConv, and Point Transformer V2 on bridge scans \cite{kellner2026semanticbridge}.
Transfer from synthetic to real bridge data has also been studied \cite{jing2022segmentation,kellner2025semisupervised}, but cross-sensor transfer between two real acquisitions has received less systematic attention.

\subsection{Domain Adaptation for Point Clouds}
Most outdoor point-cloud adaptation research targets automotive LiDAR \cite{triess2021survey}.
Input-level methods normalize sensor-specific sampling.
Complete \& Label reconstructs a canonical surface before segmentation \cite{yi2021complete}, whereas LiDAR Distillation simulates beam-induced sparsity \cite{wei2022lidar}.
Langer \etal\ render semi-synthetic scans that match a target sensor from fused source scans \cite{langer2020domain}.
Feature-level approaches use cross-modal consistency \cite{jaritz2020xmuda}, synthetic-to-real alignment \cite{xiao2022transfer}, or implicit surfaces \cite{michele2024saluda}.
Compositional semantic mix (CoSMix) exchanges semantic regions across domains \cite{saltori2023cosmix}, and ST3D stabilizes self-training with memory and curriculum mechanisms \cite{yang2021st3d}.
In remote sensing, cross-dataset airborne laser scanning (ALS) segmentation has been related to point density, intensity, and augmentation \cite{xie2021crosscity}, and SegTrans combines data selection with adversarial feature alignment for TLS-to-MLS transfer \cite{shen2023segtrans}.
Input-level methods presuppose a beam-induced sampling difference, and feature-level methods an appearance, modality, or surface-representation gap.
These methods are powerful when their assumed shift matches the data, but fixed beam geometry and vehicle viewpoint do not transfer directly to multi-view bridge acquisition.
CoSMix and ST3D rely on target pseudo-labels from the source model, whose quality depends on the target class proportions; these differ between the sensors because they observe different portions of a structure (Section~\ref{sec:coverage}).

Source-free and test-time approaches adapt normalization or predictions and assume that feature statistics or prediction confidence carry the adaptation signal.
AdaBN re-estimates batch-normalization statistics on target data \cite{li2018adabn}; TENT minimizes target prediction entropy by updating normalization affine parameters \cite{wang2021tent}; source hypothesis transfer (SHOT) combines information maximization with pseudo-labeling \cite{liang2020shot}; and efficient anti-forgetting test-time adaptation (EATA) and continual test-time adaptation (CoTTA) address forgetting and error accumulation \cite{niu2022eata,wang2022cotta}.
Class-prior shift is especially relevant because entropy minimization can reinforce majority predictions when the target prior differs from the source prior.
SHOT, EATA, and CoTTA likewise rely on confident target predictions, which can favor majority classes under such a prior change.

\subsection{Self-Training and Experimental Reliability}
Mean Teacher uses an exponential-moving-average teacher to provide consistency targets \cite{tarvainen2017mean}.
Class-balanced self-training (CBST) \cite{zou2018cbst}, confidence-regularized self-training \cite{zou2019crst}, and FixMatch \cite{sohn2020fixmatch} emphasize selective pseudo-labeling.
Selection is necessary because confirmation bias can make increasingly confident models reinforce their own errors \cite{arazo2020pseudo}.

The reliability of small benchmark gains has been questioned across machine learning \cite{henderson2018deep,melis2018state,lucic2018gans,musgrave2020metric}.
Repeated runs, controlled budgets, and score distributions are essential \cite{bouthillier2021variance,reimers2017reporting}.
Training recipes alone can rival nominal architectural gains \cite{bello2021revisiting,wightman2021resnet}, while label errors can alter rankings \cite{northcutt2021pervasive}.
The present work instantiates these concerns for cross-sensor bridge segmentation.

\section{Experimental Setup}
\subsection{Dataset and Benchmark Split}
We use the real-acquisition portion of SemanticBridge \cite{kellner2026semanticbridge}.
It contains 20 bridges from the United Kingdom and Germany, all acquired by stationary TLS systems (FARO Focus 3D X330 or Leica RTC360).
Seven bridges additionally have handheld MLS acquisitions from a Leica BLK2GO.
Labels cover nine classes: \textit{unlabeled}, \textit{ground}, \textit{high vegetation}, \textit{abutment}, \textit{superstructure}, \textit{top surface}, \textit{railing}, \textit{traffic sign}, and \textit{pillar}.
MLS labels were transferred from TLS annotations by interpolation.

We follow the benchmark's official split without excluding any scene; hereafter, we refer to it as the benchmark split:
\begin{table}[!tbp]
\caption{SemanticBridge Benchmark Split Used in This Study}
\label{tab:split}
\centering
\begin{tabular}{lll}
\toprule
Role & Partition & Bridges \\
\midrule
Source training & \texttt{tls/train} & 15 labeled \\
Source validation & \texttt{tls/val} & 5 labeled \\
Target training & \texttt{mls/train} & 4 unlabeled \\
Target evaluation & \texttt{mls/val} & 3 labeled \\
\bottomrule
\end{tabular}
\end{table}
The target evaluation scenes are bridges 13, 17, and 19.
Their TLS counterparts permit the same paired-bridge comparison used in the benchmark paper.
Bridge 19 has a registration defect discussed in Section~\ref{sec:dataquality}; it is retained to preserve protocol comparability.

\subsection{Backbone and Input Representation}
The backbone is a sparse-convolution (SpConv) U-Net corresponding to the benchmark's UNet3D family.
It has four encoder levels with 32, 64, 128, and 256 channels and 29.2 million parameters.
Points are quantized into 8-cm voxels.
The network receives discrete voxel coordinates concatenated with RGB, not continuous point coordinates.
Sub-voxel geometric variation can therefore affect the network only if it changes voxel assignment or occupancy.

For input ablations, we compute planarity, linearity, and the absolute vertical component of the normal using local principal-component analysis with $k=30$.
The descriptors are either substituted for RGB or appended to it.
All components lie in $[0,1]$ and are invariant to the rotation, reflection, and global scaling used in training.

\subsection{Training Schedules, Metrics, and Model Selection}
We use Adam with an initial learning rate (LR) of $10^{-3}$, batch size two, and 6-m cubic subclouds.
Two exponential-decay schedules isolate training-budget effects:
\begin{table}[!tbp]
\caption{Training Schedules}
\label{tab:schedules}
\centering
\begin{tabular}{lccc}
\toprule
Schedule & Epochs & $\gamma$ & Final LR multiplier \\
\midrule
Compressed & 100 & 0.955 & 0.0105 \\
Converged & 300 & 0.985 & 0.0109 \\
\bottomrule
\end{tabular}
\end{table}
Although their final multipliers are nearly identical, their learning-rate trajectories differ substantially.

We report nine-class mIoU, mIoU excluding \textit{unlabeled}, and overall accuracy (OA).
No target labels are used for checkpoint selection.
The primary checkpoint is selected by source-validation mIoU.
To characterize the converged state with lower evaluation noise, we also report the mean over evaluation points in the final 50 epochs.
Target-trajectory maxima are used only for post-hoc analysis and never as deployable model-selection results.

All experiments were performed on one NVIDIA RTX 4060 Ti with 16~GB memory. TensorACO shows, in a different problem domain, that reformulating sequential state updates as tensorized batched operations can substantially improve graphics processing unit (GPU) utilization \cite{yang2024tensorized}.
A compressed run takes approximately 2 GPU-hours and a converged run approximately 6 GPU-hours.
Twenty-nine training runs, epoch-wise metrics, configurations, and figure-generation scripts are included in the research artifact.

\section{Analysis of Cross-Sensor Differences}
\label{sec:decomposition}
\begin{figure}[!tbp]
\centering
\includegraphics[width=\textwidth]{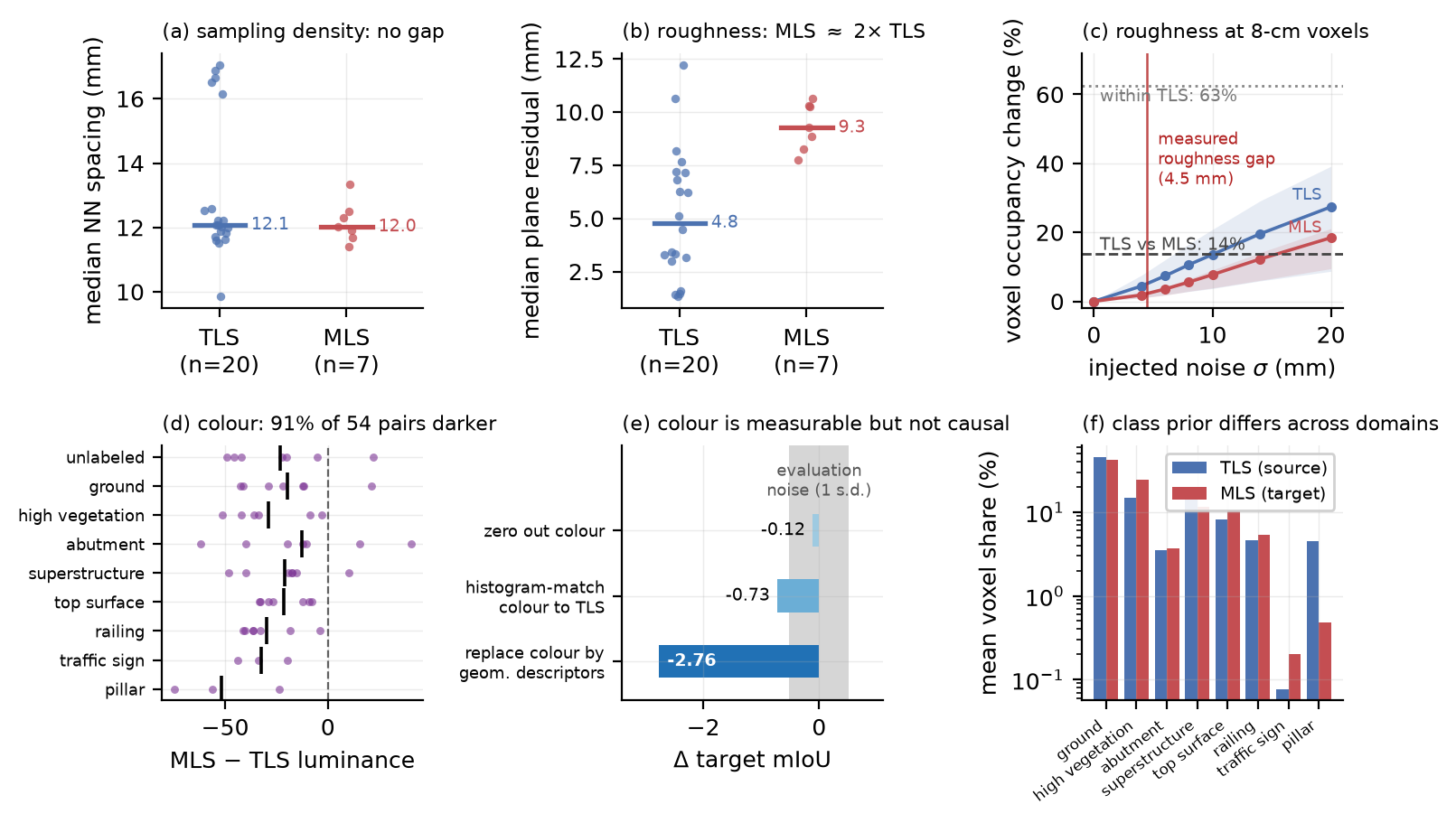}
\caption{Comparison of TLS and MLS point clouds. The panels summarize sampling density, surface roughness and voxel occupancy, luminance, and class proportions.}
\label{fig:domaingap}
\end{figure}

The comparison includes raw-point statistics and the corresponding changes in the voxelized model input.
Controlled perturbations are used to determine whether differences measured in the point clouds remain appreciable after voxelization.
Fig.~\ref{fig:domaingap} provides an overview.

\subsection{Density and Roughness}
Table~\ref{tab:geometrygap} reports raw-point and voxel statistics.
Median nearest-neighbor spacing differs by only 0.06~mm, and the median point count per 8-cm voxel differs by 3.3\%.
All three density differences are below 7\%, whereas bridge-to-bridge ranges within a domain span factors of 1.7--2.5.
The released point clouds have already been downsampled to 1~cm, largely removing the intuitive density difference between stationary and mobile raw scans.

\begin{table}[!tbp]
\caption{Density and Surface Statistics Across Domains}
\label{tab:geometrygap}
\centering
\begin{tabular}{lcc}
\toprule
Statistic & TLS (20 bridges) & MLS (7 bridges) \\
\midrule
Median nearest-neighbor spacing & 12.08 mm (9.85--17.06) & 12.02 mm (11.40--13.33) \\
Median points within 5 cm & 31 (16--40) & 29 (15--47) \\
Median points per 8-cm voxel & 10.44 (7.63--17.88) & 10.78 (5.77--15.51) \\
Median plane residual ($k=30$) & 4.79 mm (1.32--12.23) & 9.26 mm (7.77--10.64) \\
Median surface variation & 0.015 & 0.069 \\
\bottomrule
\end{tabular}
\end{table}

Surface quality does differ: the MLS median plane residual is 1.9 times the TLS value, and surface variation is 4.6 times higher.
The absolute residual difference is nevertheless about 4.5~mm, only 5.6\% of the 80-mm voxel size.
To translate this difference into the model's representation, we inject anisotropic Gaussian noise and measure the relative change in occupied voxels within a $3\times3\times3$ neighborhood.
A 4-mm perturbation, comparable to the measured roughness gap, changes occupancy by 4.4\%.
The observed TLS-to-MLS occupancy difference is 14\%, while the TLS bridge-to-bridge range is 63\% (10.50--17.06 occupied neighbors).
Perturbations of 10 and 20~mm produce changes of 13.6\% and 27.4\%, respectively.

Under the released preprocessing, density differences are small and the measured roughness difference is only weakly expressed in an 8-cm voxel representation.
This result is specific to the chosen resolution; finer voxel grids may retain more of the surface variation.

\subsection{Photometric Shift}
For well-registered bridge pairs, we compare luminance distributions within the same semantic class.
Of 54 bridge--class pairs, 90.7\% are darker in MLS; the mean luminance difference is $-24.5$ on the 0--255 scale, and the mean Wasserstein distance is 29.05.
By marginal distribution distance alone, color would appear to be a substantial source of domain difference.

The input ablation yields a different result.
We evaluated the converged RGB checkpoint (Section~\ref{sec:converged}) on identical voxels with unchanged color, with all color channels zeroed, and with the first and second MLS color moments matched to TLS (Table~\ref{tab:colorprobe}).
Zeroing color changes target mIoU by $+0.02$ and moment matching by $0.00$ relative to the unchanged baseline of 57.62.
Thus, the evaluated model makes little use of color.

\begin{table}[!tbp]
\caption{Paired Color Interventions on Identical Voxels (Nine-Class mIoU, \%)}
\label{tab:colorprobe}
\centering
\begin{tabular}{llccc}
\toprule
Checkpoint & Domain & Unchanged & Color zeroed & Moments matched \\
\midrule
Converged, benchmark aug. & MLS & 57.62 & 57.64 ($+0.02$) & 57.62 ($0.00$) \\
 & TLS & 66.57 & 66.56 ($-0.01$) & 66.57 ($0.00$) \\
100 ep., corrected aug. & MLS & 52.34 & 52.40 ($+0.06$) & 52.32 ($-0.02$) \\
 & TLS & 61.94 & 61.93 ($-0.01$) & 61.94 ($0.00$) \\
\bottomrule
\end{tabular}
\end{table}

The benchmark pipeline contains an implementation detail relevant to this result.
The reference implementation's ChromaticColorTranslation (ratio=0.01) applies a displacement of $\pm2.55$, designed for 0--255 values, to features that have already been divided by 256.
Its effective displacement is consequently comparable to the complete normalized color range.
We therefore retrained the 100-epoch RGB source-only model (seed 0) with the displacement corrected to $\pm0.01$ and all other settings unchanged.
The corrected model reaches 52.34 target mIoU at its source-selected checkpoint, compared with 53.35 for the same seed and 52.16 for the three-seed mean with the benchmark augmentation; both differences lie within the 1.88-point seed standard deviation (Section~\ref{sec:evalnoise}).
On this model, zeroing color changes target mIoU by $+0.06$ and moment matching by $-0.02$.
Color therefore remains of little use to the model after the augmentation scale is corrected.
The RGB-only results in the remaining sections use the benchmark augmentation for comparability.

\subsection{Coverage Asymmetry and Class-Prior Shift}
\label{sec:coverage}
We next compare class-specific voxel shares within the common bounding box of each registered TLS/MLS pair, using the same 8-cm grid.
Bridges 18 and 19 are excluded from this diagnostic because of registration defects, leaving five pairs.
The median MLS-to-TLS voxel-share ratios are 1.28 for abutments, 1.24 for top surfaces, and 0.68 for high vegetation (Table~\ref{tab:coverage}).

\begin{table}[!tbp]
\caption{MLS-to-TLS Voxel-Share Ratios on Five Registered Pairs}
\label{tab:coverage}
\centering
\begin{tabular}{lcc}
\toprule
Class & Median ratio & Range \\
\midrule
Abutment & 1.28 & 1.00--2.21 \\
Top surface & 1.24 & 1.11--2.10 \\
High vegetation & 0.68 & 0.58--1.16 \\
Ground & 0.90 & 0.77--1.11 \\
\bottomrule
\end{tabular}
\end{table}

\begin{figure}[!tbp]
\centering
\includegraphics[width=\textwidth]{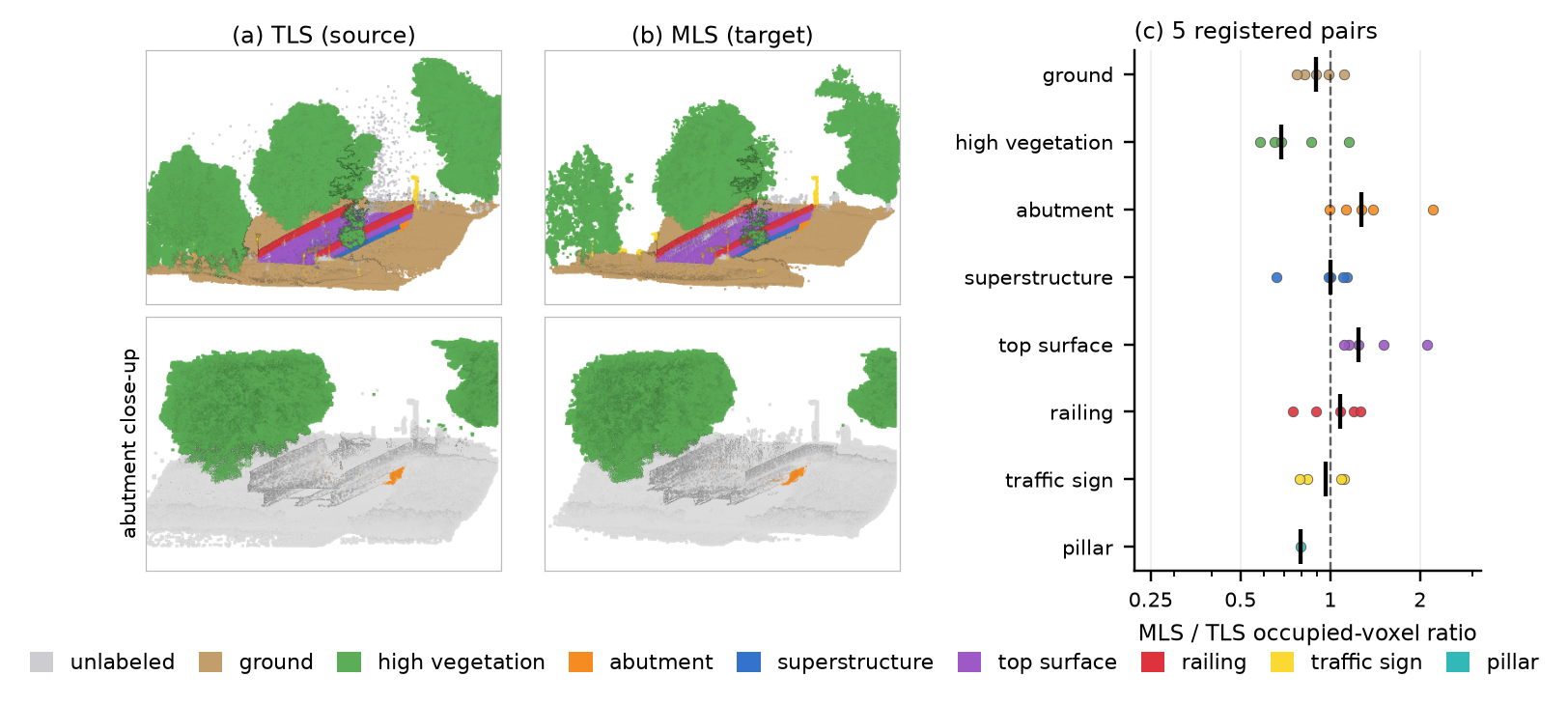}
\caption{Paired TLS/MLS coverage comparison. The semantic views and abutment close-ups illustrate viewpoint-dependent visibility, while the right panel reports class-specific occupied-voxel ratios for the five well-registered pairs. MLS provides more complete abutment and top-surface coverage, whereas TLS observes more high vegetation.}
\label{fig:coveragepair}
\end{figure}

The observed ratios are consistent with the different sensor viewpoints.
A mobile scanner passing under or along a bridge observes abutments and deck surfaces more completely, whereas an elevated stationary scanner covers more distant vegetation.
The paired visualization and across-pair ratios in Fig.~\ref{fig:coveragepair} show both effects directly.
Coverage asymmetry changes class priors even when point density is matched.
The same components account for much of the improvement obtained with the longer training schedule.

\subsection{Input-Modality Ablation}
The three geometric descriptors have a mean cross-domain Wasserstein distance of 0.030 on a normalized scale, compared with 0.114 for luminance; they are approximately 3.9 times more stable across sensors.
The smaller distributional difference does not translate into higher segmentation accuracy.
Replacing color with the descriptors reduces target mIoU by 2.76 under the compressed schedule.
Appending descriptors to RGB reduces it by 1.91 under that schedule and by approximately 0.3--0.5 in converged-state comparisons.
Voxel coordinates already encode substantial geometry, making the descriptors partly redundant, whereas color can supply orthogonal information even if the trained model uses it weakly.

\subsection{Data-Quality Bounds}
\label{sec:dataquality}
Because MLS labels were interpolated from TLS annotations, the two label sets are not identical.
After projection to a common point set with a 10-cm matching threshold, label agreement is 96.9\% and aggregate mIoU excluding \textit{unlabeled} is 88.5\%.
Per-bridge mIoU excluding \textit{unlabeled} ranges from 48.7\% to 95.3\%; abutment intersection over union (IoU) is 69.3\% and railing IoU is 81.1\%, while ground and high vegetation exceed 97\%.
These values characterize the consistency of the released annotations rather than a theoretical upper bound, and estimates for individual bridges remain sensitive to low point-match rates.

\begin{figure}[!tbp]
\centering
\includegraphics[width=0.62\textwidth]{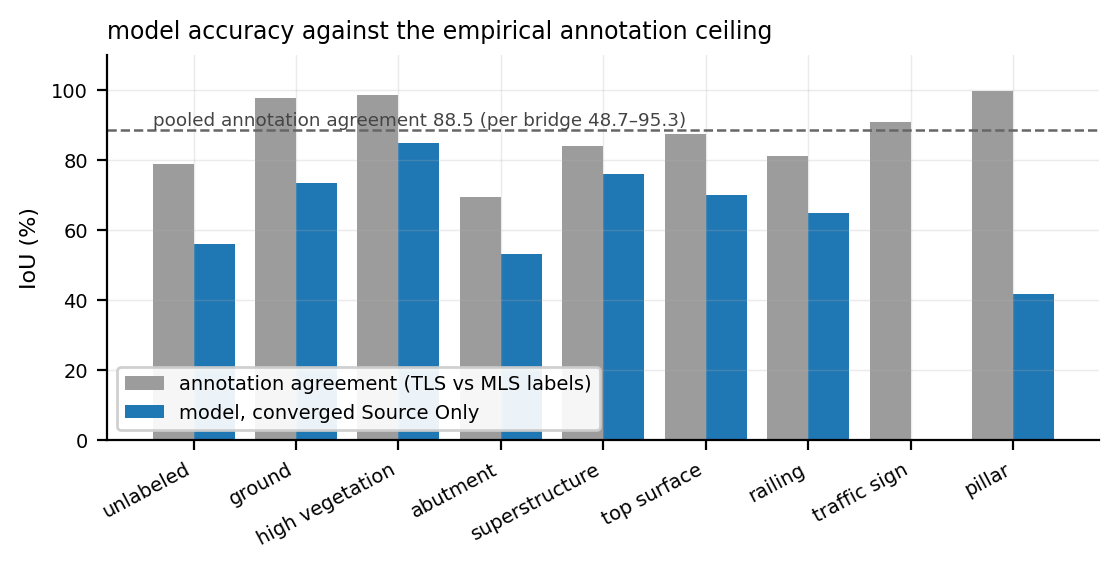}
\caption{Per-class cross-annotation agreement compared with the converged source-only model. Abutments and railings have lower TLS--MLS label agreement than the dominant background classes.}
\label{fig:annotation}
\end{figure}

Figure~\ref{fig:annotation} places the model's per-class IoU beside this empirical annotation agreement, highlighting that several difficult structural classes also have less consistent transferred labels.

\begin{figure}[!tbp]
\centering
\includegraphics[width=0.62\textwidth]{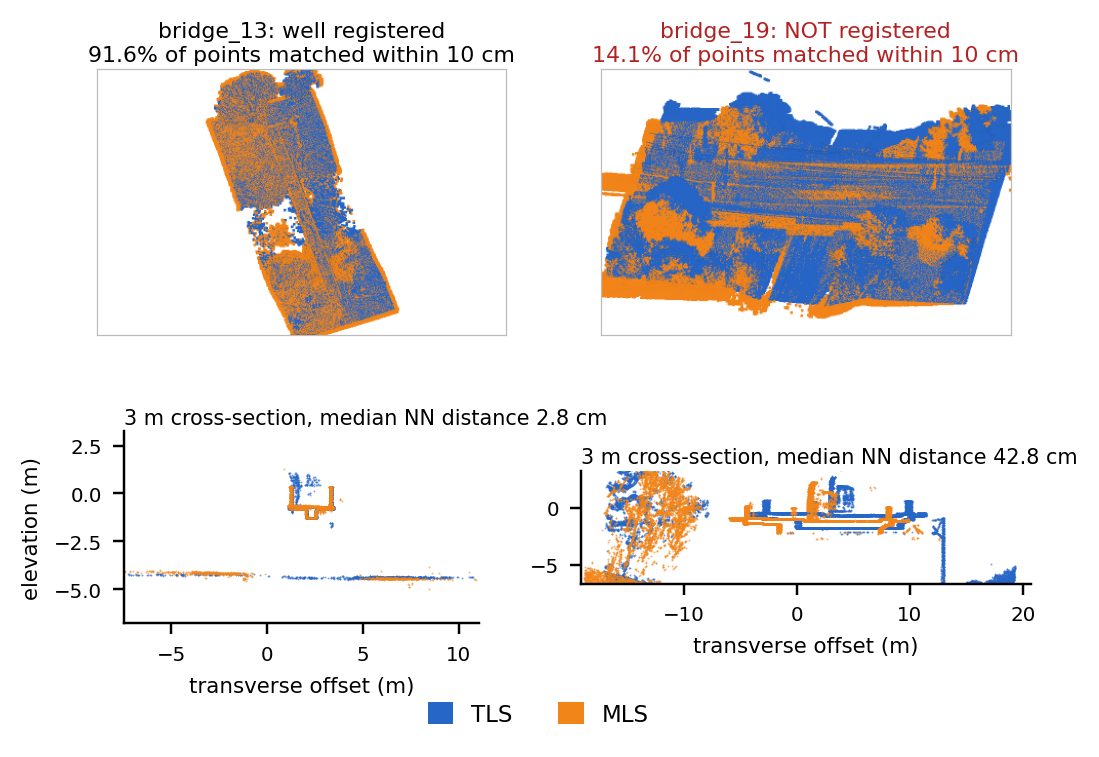}
\caption{Examples of a well-registered pair (bridge 13) and a defective pair (bridge 19). Bridge 19 exhibits visible lateral and vertical offsets and remains in the benchmark target-evaluation split.}
\label{fig:registration}
\end{figure}

Two released pairs exhibit marked registration offsets.
Bridge 18 has a 5.8\% nearest-neighbor match rate within 10~cm and a median nearest-neighbor distance of 107~cm.
Bridge 19 has a 14.1\% match rate and a median distance of 43~cm, compared with 2.6--14.5~cm medians for the five better-registered pairs.
Bridge 19 is part of the benchmark evaluation set and has only 23.3\% cross-annotation mIoU.
We retain it to match the benchmark split; its MLS mIoU is the highest of the three paired evaluation bridges, and that of the well-registered bridge 13 is the lowest (Table~\ref{tab:perbridge}).

\section{Baseline and Adaptation Results}
\subsection{Converged Source-Only Baseline}
\label{sec:converged}
Table~\ref{tab:converged} reports the 300-epoch runs.
The simplest RGB source-only configuration produces the strongest converged-state estimate: 57.72 target mIoU at the source-selected checkpoint and 58.41 averaged over evaluation points in the final 50 epochs.
The latter summarizes late-stage performance and does not correspond to a single released checkpoint.

\begin{table}[!tbp]
\caption{Results Under the 300-Epoch Converged Schedule (Single Seed, \%)}
\label{tab:converged}
\centering
\resizebox{\textwidth}{!}{%
\begin{tabular}{lccccc}
\toprule
Configuration & Source mIoU & Target, source-selected & Target, final-50 mean & Target peak$^\dagger$ & OA \\
\midrule
Source only, RGB & 66.09 & 57.72 (ep.~99) & \textbf{58.41} & 59.13 (ep.~274) & 87.44 \\
Source only, RGB+geometry & 67.41 & 58.41 (ep.~269) & 58.09 & 58.70 (ep.~259) & 87.45 \\
Filtered self-training from converged model & 68.63 & 57.44 (ep.~79) & 55.65 & 58.22 (ep.~44) & 87.67 \\
\bottomrule
\multicolumn{6}{l}{\footnotesize $^\dagger$Target-label-dependent diagnostic only; not used for model selection.}
\end{tabular}}
\end{table}

To compare fairly with SemanticBridge, we re-evaluate the same model on the three bridges for which both TLS and MLS appear in the benchmark's paired protocol.
Table~\ref{tab:benchmark} separates absolute target accuracy from the cross-sensor change.

\begin{table}[!tbp]
\caption{Paired-Bridge Comparison With SemanticBridge (\%)}
\label{tab:benchmark}
\centering
\begin{tabular}{lccc}
\toprule
Model & TLS & MLS & Change \\
\midrule
UNet3D, reported \cite{kellner2026semanticbridge} & 68.2 & 57.4 & $-10.8$ \\
KPConv, reported \cite{kellner2026semanticbridge} & 70.9 & \textbf{59.5} & $-11.4$ \\
PTv2, reported \cite{kellner2026semanticbridge} & 64.4 & 57.5 & $-6.9$ \\
SpConv U-Net, 100 epochs & 58.50 & 53.30 & $-5.21$ \\
SpConv U-Net, 300 epochs & 62.39 & 57.80 & $\mathbf{-4.59}$ \\
\bottomrule
\end{tabular}
\end{table}

\begin{figure}[!tbp]
\centering
\includegraphics[width=0.55\textwidth]{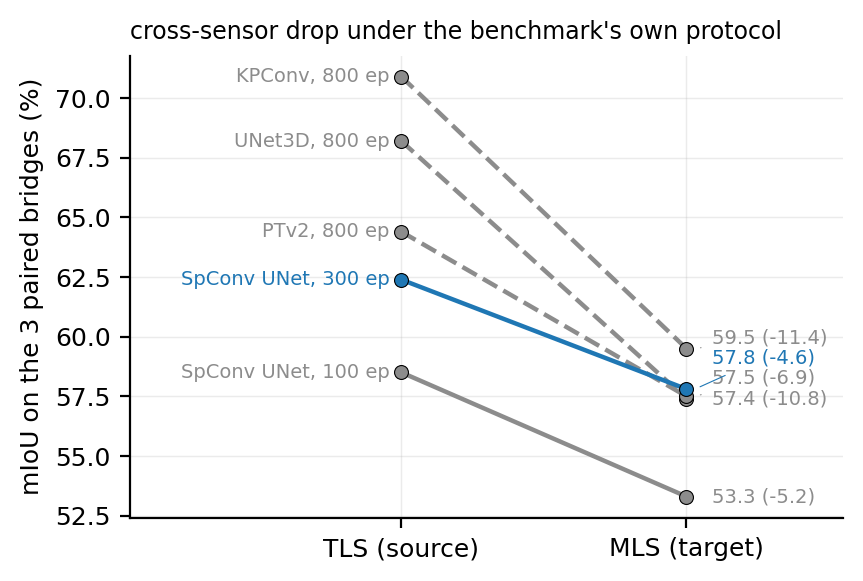}
\caption{Source-to-target slopes under the paired-bridge benchmark protocol. The converged baseline has a smaller drop, but its lower TLS score confounds interpretation of the drop magnitude.}
\label{fig:benchmark}
\end{figure}

The converged model's MLS mIoU is comparable to the reported UNet3D and Point Transformer V2 values and below KPConv's 59.5.
Its 4.59-point drop is smaller than the reported 6.9--11.4 range, but this does not by itself demonstrate superior invariance: its TLS mIoU is also lower than all three reported source scores, mechanically limiting the possible drop.
One identifiable discrepancy is \textit{traffic sign}, whose paired-TLS IoU is zero in our runs but 0.196 for the reported UNet3D, accounting for approximately 2.2 source-mIoU points.
The benchmark models were also trained for 800 epochs.
The converged implementation therefore reaches a target score similar to those reported for UNet3D and Point Transformer V2, although differences in source accuracy and training setup prevent a direct comparison of cross-sensor robustness.

\begin{table}[!tbp]
\caption{Per-Bridge Scores of the Converged RGB Source-Only Model (Whole-Cloud Majority Vote, \%)}
\label{tab:perbridge}
\centering
\begin{tabular}{lccccl}
\toprule
Bridge & TLS & MLS & Change & Match rate$^\ddagger$ & Registration \\
\midrule
13 & 38.26 & 29.15 & $-9.11$ & 91.6 & good \\
17 & 64.69 & 58.69 & $-6.00$ & 40.6 & good \\
19 & 63.97 & 62.91 & $-1.06$ & 14.1 & poor$^{\S}$ \\
\midrule
Pooled & 62.39 & 57.80 & $-4.59$ & & \\
\bottomrule
\multicolumn{6}{p{0.6\textwidth}}{\footnotesize $^\ddagger$Nearest-neighbor TLS--MLS match rate within 10~cm. $^{\S}$Cross-annotation mIoU 23.3\%.}
\end{tabular}
\end{table}

Table~\ref{tab:perbridge} reports per-bridge scores of the 300-epoch model.
The pooled MLS mIoU of 57.80 falls to 53.39 without bridge 19 and rises to 61.04 without bridge 13.

\begin{table}[!tbp]
\caption{Test-Time Adaptation From the Converged RGB Checkpoint (Paired Evaluation, Nine-Class mIoU, \%)}
\label{tab:convtta}
\centering
\begin{tabular}{lcccccc}
\toprule
 & \multicolumn{3}{c}{Target (MLS)} & \multicolumn{2}{c}{Source (TLS)} & \\
\cmidrule(lr){2-4}\cmidrule(lr){5-6}
Method & Before & After & Change & Before & After & Time \\
\midrule
AdaBN \cite{li2018adabn} & 57.62 & 58.07 & $+0.45$ & 66.57 & 66.51 & 9.9 s \\
TENT \cite{wang2021tent} & 57.62 & 54.30 & $-3.32$ & 66.57 & 62.29 & 9.2 s \\
\bottomrule
\end{tabular}
\end{table}

We also applied AdaBN and TENT to the converged RGB checkpoint (Table~\ref{tab:convtta}).
Both methods adapt on the unlabeled \textit{mls/train} bridges in one pass of 383 batches, and target and source mIoU are measured before and after adaptation on the same cached validation voxels (403 MLS and 600 TLS batches).
The unadapted value of 57.62 differs from the 57.72 in Table~\ref{tab:converged} because the cached voxels differ from those sampled during training-time evaluation.
The $+0.45$ change of AdaBN lies within the 0.27--0.64 standard deviation of repeated evaluation (Section~\ref{sec:evalnoise}).
TENT updates only the affine parameters of the normalization layers (70 parameter tensors) by entropy minimization with a learning rate of $10^{-3}$.
Its 3.32-point target decrease exceeds the largest within-checkpoint evaluation range (2.03) but is smaller than the 12.19-point decrease under the compressed schedule (Table~\ref{tab:adaptation}).

\subsection{Published Adaptation Methods Under the Compressed Schedule}
Table~\ref{tab:adaptation} summarizes adaptation under the 100-epoch schedule.
\begin{table}[!tbp]
\caption{Adaptation Results Under the 100-Epoch Compressed Schedule (\%)}
\label{tab:adaptation}
\centering
\resizebox{\textwidth}{!}{%
\begin{tabular}{llccccc}
\toprule
Method & Features & Seeds & Source mIoU & Target mIoU & \shortstack{Vs. common baseline$^{a}$\\(reference)} & \shortstack{Paired component\\effect$^{b}$} \\
\midrule
Source only & RGB & 0--2 & $62.60\pm0.39$ & $52.16\pm1.88$ & --- & --- \\
Source only & RGB+geometry & 0--1 & 61.21 & 51.34 & $-1.91$ (53.25) & --- \\
Source only & Geometry only & 0 & 61.44 & 50.59 & $-2.76$ (53.35) & --- \\
AdaBN \cite{li2018adabn} & RGB & 0 & 61.23 & 51.68 & $-1.67$ (53.35) & $-1.67$ \\
TENT \cite{wang2021tent} & RGB & 0 & 49.14 & 41.16 & $-12.19$ (53.35) & $-12.19$ \\
Mean Teacher \cite{tarvainen2017mean} & RGB+geometry & 0 & 63.23 & 53.01 & $-0.34$ (53.35) & $+0.79$ \\
Class-balanced thresholding \cite{zou2018cbst} & RGB & 0 & 64.82 & 55.06 & $+1.71$ (53.35) & $+1.71$ \\
Confidence-filtered self-training & RGB & 0--1 & 64.63 & 55.16 & $+1.92$ (53.25) & $+1.92$ \\
Confidence-filtered self-training & RGB+geometry & 0--1 & 65.21 & 55.85 & $+2.61$ (53.25) & $+4.51$ \\
Filtered self-training + geometric gate & RGB+geometry & 0--1 & 65.28 & \textbf{56.48} & $\mathbf{+3.24}$ (53.25) & $\mathbf{+5.14}$ \\
\bottomrule
\end{tabular}}
\par\vspace{3pt}\parbox{\textwidth}{\footnotesize The first row gives the mean $\pm$ standard deviation over seeds 0--2; other multi-seed rows give means over seeds 0--1 (e.g., the gated row: 56.59 for seed 0 and 56.37 for seed 1).
$^{a}$Target mIoU minus the RGB source-only target mIoU of the matching seeds, shown in parentheses: 53.35 (seed 0) for one-seed rows and 53.25 (mean of seeds 0 and 1) for two-seed rows. The three-seed mean of 52.16 is not used as a subtrahend.
$^{b}$Target mIoU minus that of the method's own warm start: the seed-0 RGB model (53.35) for AdaBN, TENT, and class-balanced thresholding; the seeds 0--1 RGB models (53.25) for RGB self-training; the seed-0 RGB+geometry model (52.22) for Mean Teacher; and the seed-matched RGB+geometry models (52.22 and 50.46, averaged over seeds) for the two RGB+geometry self-training rows.}

\end{table}
AdaBN decreases target mIoU by 1.67 despite exact cumulative estimation of target normalization statistics.
This suggests that re-estimating low-order batch-normalization moments does not address the relevant differences between the two domains.
TENT decreases mIoU by 12.19.
TENT was developed primarily for distribution corruptions, and its entropy objective can reinforce majority-class predictions when class proportions change; the result is therefore specific to this setting.
Mean Teacher gains only 0.79 over its own warm start and reaches its best value at epoch 3.
It shares the teacher, exponential-moving-average update, and warm start with filtered self-training; the principal difference is its 100\% pseudo-label acceptance.
Adding a 0.9 confidence threshold improves the corresponding result, suggesting that pseudo-label selection is important in this experiment.

\subsection{Self-Training Before and After Convergence}
Under the compressed schedule, confidence-filtered self-training with geometric gating is 3.24 points above the same-budget, seed-matched RGB baseline (53.25; Table~\ref{tab:adaptation}).
Its largest class gain is on abutments (46.49 to 57.71), a class whose relative coverage differs between TLS and MLS.
When self-training begins from a converged RGB+geometry checkpoint, however, this improvement is no longer observed.
Relative to the warm start, target mIoU changes by $-0.97$ using source-selected checkpoints and by $-2.44$ using final-50-epoch means.
Even the target-dependent trajectory peak, 58.22 at epoch 44, remains below the warm-start convergence estimate.

\begin{figure}[!tbp]
\centering
\includegraphics[width=0.82\textwidth]{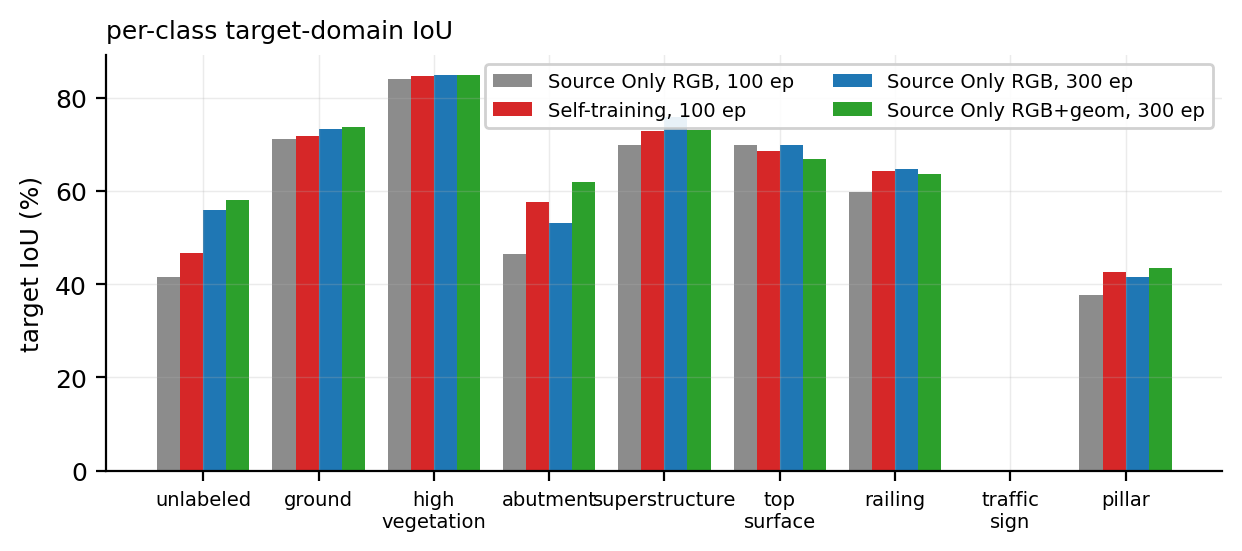}
\caption{Target-domain per-class IoU for representative compressed- and converged-schedule runs. Longer source training produces gains comparable to or larger than compressed-schedule self-training for several structural classes; traffic-sign IoU is zero in all four runs.}
\label{fig:perclass}
\end{figure}

The per-class comparison in Fig.~\ref{fig:perclass} shows that the improvement is concentrated in particular components rather than distributed uniformly across classes.

\begin{figure}[!tbp]
\centering
\includegraphics[width=0.55\textwidth]{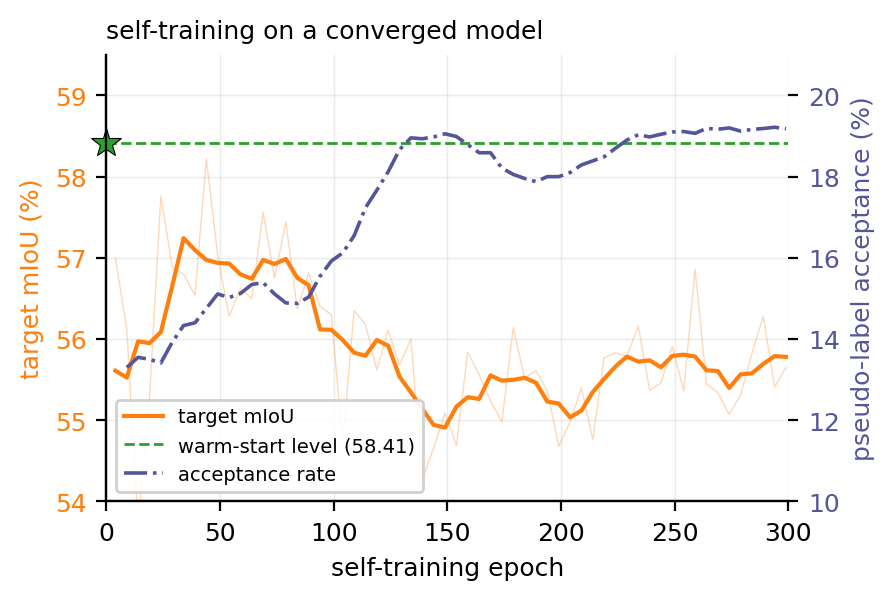}
\caption{Confidence-filtered self-training initialized from a converged model. Target accuracy never recovers its initial level, while pseudo-label acceptance rises from 11.7\% (epoch 5) to 19.0\% (epoch 299), consistent with confirmation bias. The acceptance curve is a 9-point moving average.}
\label{fig:selftrain}
\end{figure}

The pseudo-label acceptance rate rises from 11.7\% at epoch 5 to 19.0\% at epoch 299 while target performance declines.
This trend is compatible with confirmation bias, although pseudo-label accuracy was not measured directly.
A comparison at a similar computational cost produces the same ordering.
The compressed self-training run, including its 81-epoch warm start, consumes 112 epochs and reaches 56.59, whereas the source-only converged schedule reaches 57.72 by epoch 114.
Other self-training designs, including stronger class reweighting, confidence regularization, or cross-domain mixing, may behave differently.
For the configuration considered here, standard thresholded self-training does not improve the converged model.

\section{Effects of Training and Evaluation}
\subsection{Stochastic Evaluation}
\label{sec:evalnoise}
Evaluation samples random subclouds and randomly selects representative points within voxels.
The same fixed checkpoint therefore produces different values across evaluations; the observed validation subcloud count ranges from 771 to 809.
We evaluate ten checkpoints eight times each.
Target mIoU standard deviations range from 0.268 to 0.642, with a mean of 0.512, and within-checkpoint ranges span 0.82--2.03 points.
The standard error after eight evaluations is approximately 0.18.

\begin{figure}[!tbp]
\centering
\includegraphics[width=0.55\textwidth]{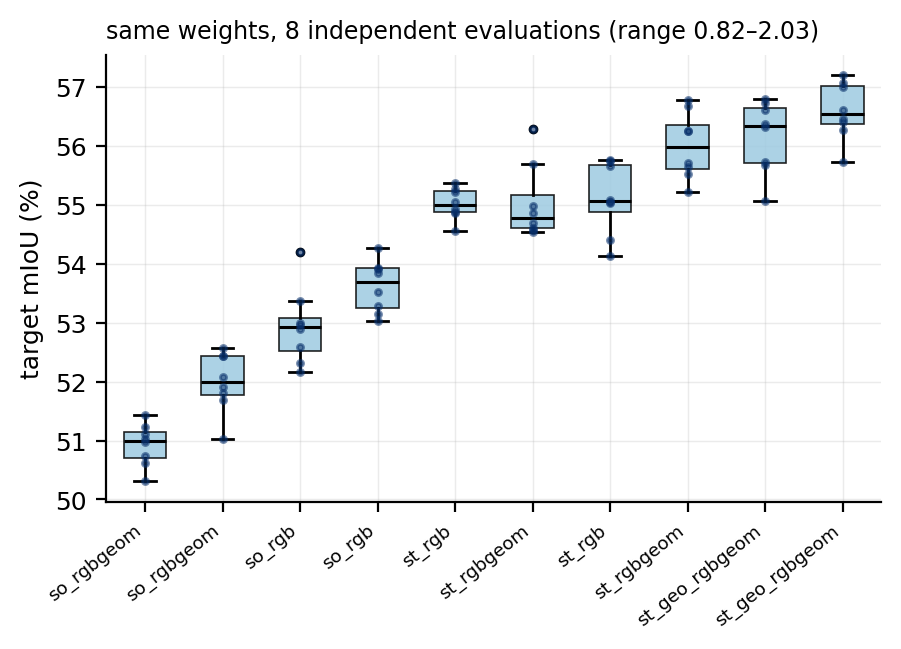}
\caption{Eight repeated evaluations of ten fixed checkpoints. Protocol stochasticity is comparable to many reported component effects.}
\label{fig:evalnoise}
\end{figure}

Consequently, a difference of about 0.5 points can arise from repeated evaluation of the same checkpoint.

Variation across training seeds is larger.
The three 100-epoch RGB source-only seeds yield source mIoU of 62.98, 62.63, and 62.20 ($62.60\pm0.39$) and target mIoU of 53.35, 53.14, and 50.00 ($52.16\pm1.88$).
All 300-epoch results are from seed 0.
Target-mIoU differences smaller than 1.88 points therefore lie within seed variation.

\subsection{Learning-Rate Annealing Rather Than Epoch Count}
For the same seed and RGB input, the compressed schedule yields 53.35 target mIoU, whereas the 300-epoch schedule yields 57.72 at the source-selected checkpoint and 58.41 in its final-stage mean.
The difference cannot be attributed only to the additional epochs.
At epoch 75, the compressed learning rate is 0.0316 of its initial value; at epoch 99, the converged schedule retains 0.224.
When both runs are inspected at epoch 100, the slower-decay trajectory is already approximately 3.2 points better.
The performance gap is already present before the additional 200 epochs, indicating that the learning-rate trajectory contributes substantially to the result.

\begin{figure}[!tbp]
\centering
\includegraphics[width=0.85\textwidth]{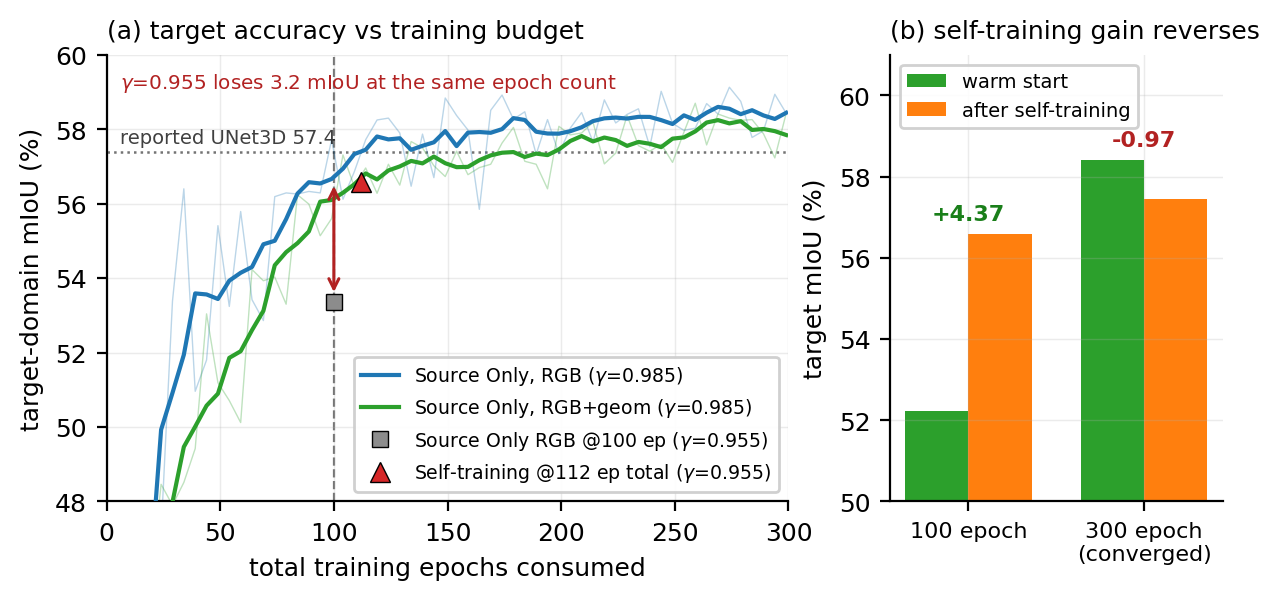}
\caption{Training-budget analysis. (a) Target mIoU versus total consumed epochs, including warm-start cost. At 100 epochs, changing only the annealing schedule produces an approximately 3.2-point difference. (b) The apparent effect of self-training changes sign after the baseline converges.}
\label{fig:budget}
\end{figure}

Matching only the final learning-rate multiplier is therefore insufficient when shortening the schedule.
In this experiment, gains measured against the compressed baseline partly reflect incomplete optimization rather than a higher final accuracy due to adaptation.

\begin{figure}[!tbp]
\centering
\includegraphics[width=\textwidth]{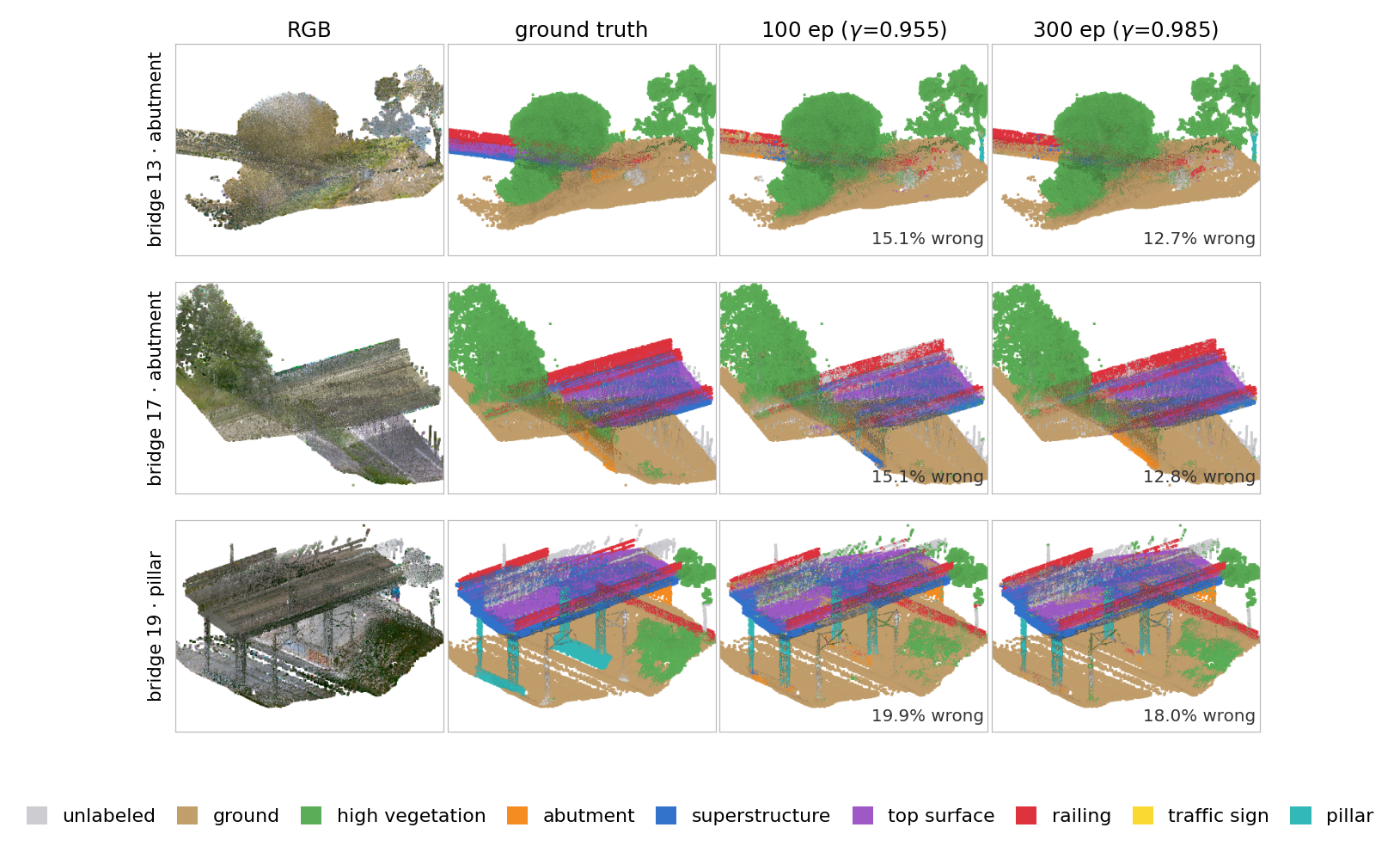}
\caption{Local qualitative comparison of the compressed and converged source-only schedules. The slower-decay 300-epoch schedule reduces errors around abutments on bridges 13 and 17 and around pillars on bridge 19. Percentages report the fraction of incorrectly labeled points in each displayed crop.}
\label{fig:budgetqual}
\end{figure}

The local comparisons in Fig.~\ref{fig:budgetqual} are consistent with the quantitative trend: longer, slower-decay training mainly repairs predictions on structural components that are incompletely learned under the compressed schedule.

\section{Discussion}
\subsection{Implications for Cross-Sensor Adaptation}
Automotive LiDAR adaptation often leverages fixed scan lines, vehicle viewpoint, and temporal acquisition \cite{triess2021survey,yi2021complete,wei2022lidar}.
Bridge scans lack a common spherical projection or trajectory.
In SemanticBridge, the released 1-cm preprocessing further removes the expected density mismatch.
At an 8-cm voxel scale, roughness is weakly resolved, and the trained models make little use of color with either the benchmark or the scale-corrected color augmentation.
Methods targeting these variables are therefore unlikely to yield large gains with the present preprocessing and voxel resolution.

Coverage differs from the other factors because a change in viewpoint can remove semantic evidence altogether.
This observation motivates training-time simulation of target-like occlusion, explicit correction for class-proportion changes, and structural constraints for partially observed bridge components.
Their effects would need to be separated from those of source-model optimization and evaluation variability.

Beyond changing voxel resolution, a promising direction is to learn a compact representation of local 3-D structure.
In medical action evaluation, encoding joint trajectories as adaptive anisotropic 3-D Gaussian tokens can preserve structured motion information with low computational cost \cite{yang2026multivariate}.
Whether such a representation improves cross-sensor segmentation remains to be evaluated.

The target-only supervised experiment illustrates the limited size and class coverage of the MLS training split.
Training on all labels from the four target-training bridges yields only 23.56 mIoU, with zero IoU for abutment, traffic sign, and pillar.
Pillars appear in only one of the four bridges and account for 1.07\% of its voxels; traffic signs account for 0.33--0.50\%.

\subsection{Limitations}
The experiments use one benchmark, one sparse-convolution backbone, and an 8-cm voxel grid. The density, roughness, and color results are therefore conditional on the SemanticBridge preprocessing and this representation resolution; other datasets, architectures, and finer voxel grids may produce different results.
Future work includes isolating viewpoint-dependent coverage with a controlled occlusion experiment, extending the converged-start comparison to the further recent methods discussed in Section~\ref{sec:related}, and applying the controlled protocol to other architectures, voxel resolutions, and datasets.
RGB-only runs use the official color translation, a displacement of $\pm2.55$ on features already divided by 256.
Runs with geometric features use a corrected displacement of $\pm0.01$ on every feature column: the official operator emits a three-dimensional vector, and a $\pm2.55$ offset would dominate the geometric channels.
Retraining the 100-epoch RGB model with the corrected displacement reaches 52.34 target mIoU (seed 0), against 53.35 for official seed 0 and $52.16\pm1.88$ for the three official seeds.
The RGB versus RGB+geometry comparison is therefore not explained by this implementation difference.

\section{Conclusion}
This study examined TLS-to-MLS transfer at the input resolution of an 8-cm sparse-voxel network.
After SemanticBridge preprocessing, TLS and MLS have similar sampling density, and the measured surface-roughness difference produces a limited change in voxel occupancy.
Color distributions differ between the sensors, but zeroing or moment-matching color changes target mIoU by less than 0.1 point with both the benchmark and the scale-corrected color augmentation.
The two acquisition systems also observe different portions of the bridge, producing appreciable differences in class proportions.

A source-only model trained with a sufficiently slow learning-rate decay reaches 57.80 mIoU on the paired-bridge MLS benchmark protocol, comparable to two published backbones and below KPConv.
Confidence-filtered self-training helps an under-converged baseline but does not improve the converged model.
Repeated evaluations and multi-seed runs show that target-domain variability is comparable to several adaptation gains reported in the experiments.
These results emphasize the need to account for source-model convergence, evaluation variability, and data quality when comparing cross-sensor adaptation methods on this benchmark.

\section*{Acknowledgment}
Not applicable.

{\small
\bibliographystyle{IEEEtran}
\bibliography{ref}}

\end{document}